\documentclass[letterpaper, 10 pt, conference]{ieeeconf}  
\IEEEoverridecommandlockouts
\usepackage{graphicx}
\usepackage{epstopdf}
\usepackage{amsmath}
\usepackage{amssymb}
\usepackage{subfigure}
\usepackage{multirow}
\usepackage{pbox}
\usepackage{cite}
\usepackage{algorithm}
\usepackage[noend]{algpseudocode}
\usepackage{booktabs}
\usepackage{wrapfig}
\usepackage{lscape}
\usepackage{bm}
\usepackage{url}
\usepackage{esvect}
\usepackage{xcolor,soul}
\usepackage{txfonts}
\usepackage[normalem]{ulem}
\usepackage{tikz}
\usetikzlibrary{matrix,calc}

\DeclareSymbolFont{extraup}{U}{zavm}{m}{n}
\DeclareMathSymbol{\varheart}{\mathalpha}{extraup}{86}
\DeclareMathSymbol{\vardiamond}{\mathalpha}{extraup}{87}

\DeclareMathOperator*{\argmin}{arg\,min}
\algnewcommand\algorithmicforeach{\textbf{for each}}
\algdef{S}[FOR]{ForEach}[1]{\algorithmicforeach\ #1\ \algorithmicdo}

\newcommand{\pluseq}{\mathrel{+}=}
\newcommand\AP{A\!P}

\newtheorem{problem}{Problem}

\renewcommand{\baselinestretch}{0.95}

\begin{document}

\title{\LARGE \bf Robust Online Epistemic Replanning of Multi-Robot Missions}
\author{Lauren Bramblett, Branko Miloradovi\'c, Patrick Sherman, Alessandro V. Papadopoulos, and Nicola Bezzo
\thanks{Lauren Bramblett, Patrick Sherman, and Nicola Bezzo are with the Departments of Systems and Information Engineering and Electrical and Computer Engineering, University of Virginia, Charlottesville, VA 22904, USA. 
Email: {\tt \{qbr5kx, ukw4tc, nb6be\}@virginia.edu}}
\thanks{Branko Miloradovi\'c and Alessandro Vittorio Papadopoulos are with the Division of Intelligent Future Technologies, M\"alardalen University, Sweden. 
Email: {\tt \{branko.miloradovic,alessandro.papadopoulos\}@mdu.se}}}

\maketitle

\begin{abstract}
As Multi-Robot Systems (MRS) become more affordable and computing capabilities grow, they provide significant advantages for complex applications such as environmental monitoring, underwater inspections, or space exploration. However, accounting for potential communication loss or the unavailability of communication infrastructures in these application domains remains an open problem.
Much of the applicable MRS research assumes that the system can sustain communication through proximity regulations and formation control or by devising a framework for separating and adhering to a predetermined plan for extended periods of disconnection. The latter technique enables an MRS to be more efficient, but breakdowns and environmental uncertainties can have a domino effect throughout the system, particularly when the mission goal is intricate or time-sensitive. To deal with this problem, our proposed framework has two main phases: i) a centralized planner to allocate mission tasks by rewarding intermittent rendezvous between robots to mitigate the effects of the unforeseen events during mission execution, and ii) a decentralized replanning scheme leveraging epistemic planning to formalize belief propagation and a Monte Carlo tree search for policy optimization given distributed rational belief updates.  
The proposed framework outperforms a baseline heuristic and is validated using simulations and experiments with aerial vehicles. 

\vspace{3pt}
\noindent\emph{Note---}Videos are provided in the supplementary material and also at {\url{https://www.bezzorobotics.com/lb-iros24}}.
\end{abstract}
\section{Introduction}\label{sec:intro}
As robotics and artificial intelligence continue to evolve, Multi-Robot Systems (MRS) have emerged as a particularly intriguing research domain. 
At the core of MRS research lies the concept of robots working together to achieve shared objectives. This collaboration can be seen in various applications, from search and rescue operations to underwater exploration. To collaborate effectively, robots must communicate and coordinate their activities, but challenges often arise when communication is limited or compromised. Consider a scenario in which an MRS has limited communication. In a multi-robot mission, there are typically no policies in place for limited connectivity. As such, failures or disturbances can cause the entire operation to be inefficient or compromised because robots cannot adapt to new information. 

In this work, we focus on the following question: \textit{How can we ensure cooperative and efficient behavior for task allocation when a centralized predefined plan must change at runtime?} This question is an expansion of our previous work~\cite{bramblett2023epi}, allowing for the elimination of certain limiting suppositions and making the work more suitable for practical scenarios where tasks are known but unforeseen changes in the environment or MRS may occur. Our proposed solution has two main components: 1) a centralized mission planner that accounts for intermittent rendezvous, promoting the discovery of failures and inefficiencies in the MRS if something does not go according to plan, and 2) an efficient runtime plan adaptation that leverages our recent epistemic planning research~\cite{bramblett2023frontiers} to reason about the likely knowledge and intentions of others based on the current epistemic state and dynamically reassign tasks. Our proposed framework enables MRS to cooperate, given limited communication and an uncertain operating environment.



\begin{figure}[t!]
\centering
\includegraphics[width=0.47\textwidth,trim={0cm 0.0cm 0cm 0.0cm},clip]{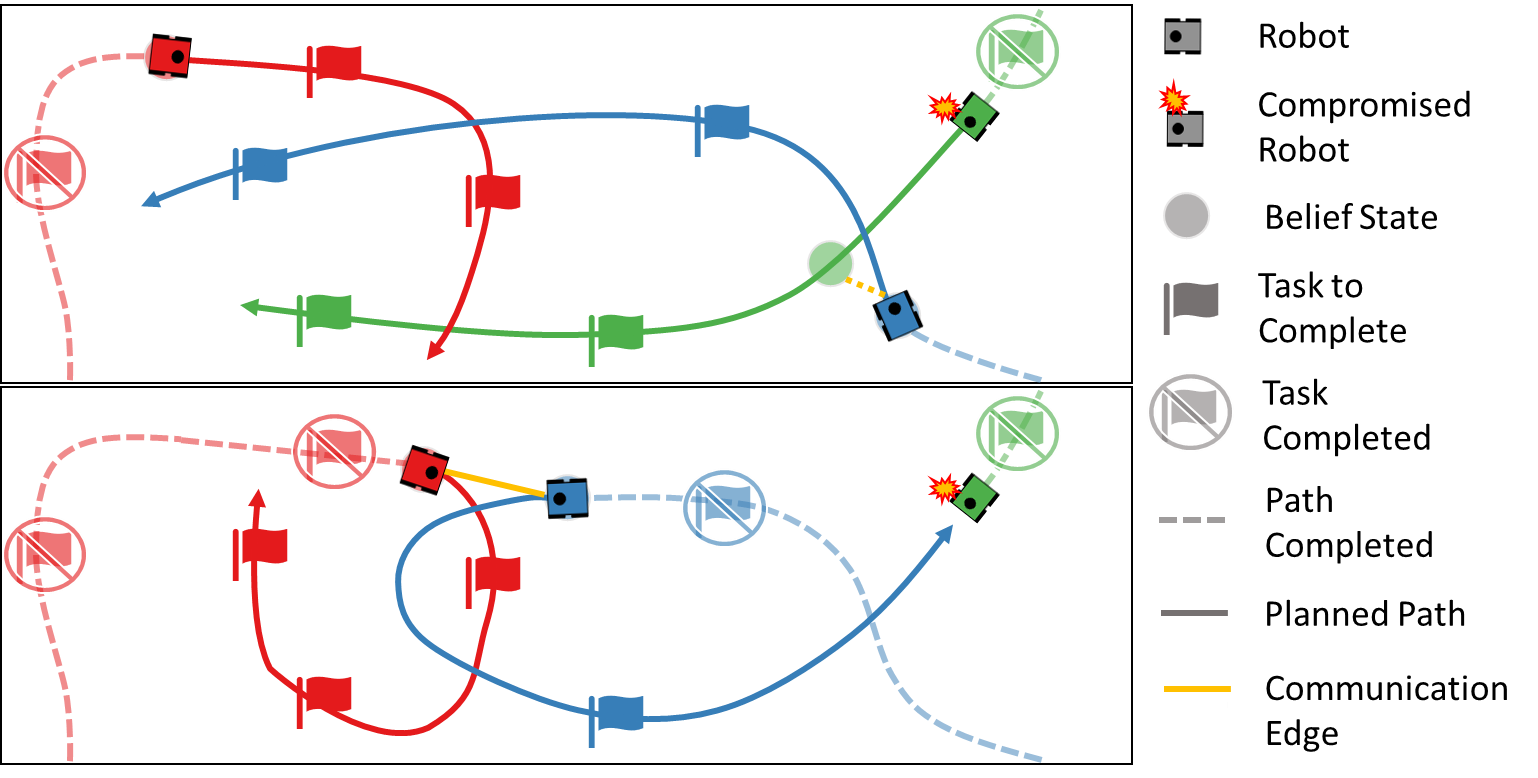}
\vspace{-10pt}
\caption{Pictorial representation of the problem presented in the paper. The green robot fails, and the blue robot observes that its belief is false. The blue robot routes to share this information with the red robot, reallocating tasks in the environment before searching for the green robot.}
\label{fig:architecture}
\vspace{-20pt}
\end{figure}

Consider the example in Fig.~\ref{fig:architecture}, where three robots complete tasks based on an initial centralized plan. During disconnection, each robot maintains a set of possible {\em belief} states for other robots and a set of {\em empathy} states that other robots might believe about it. In the top frame, the blue robot realizes that its belief of the green robot is false. It then communicates this to the red robot, and consequently, the red and blue robots reallocate their tasks (bottom frame). The blue robot is assigned to locate the green robot to ensure all tasks are completed. In this manner, robots can successively reason based on their local observations. 

The contributions of this work are two-fold: i) a genetic algorithm for multi-robot mission planning in a centralized manner, accounting for intermittent rendezvous at user-defined priorities, and ii) an epistemic planning framework for local replanning, utilizing a Monte Carlo tree search to maximize policy reward based on knowledge and beliefs about the system and environment. To the best of our knowledge, this is the first paper combining epistemic logic with runtime task allocation adaptations with intermittent communication. We show that our method outperforms a baseline heuristic in which robots complete their assigned tasks before backtracking to find faulty robots.
\section{Related Work}\label{sec:relWork}
Task allocation and planning are challenging problems that have attracted researchers from different disciplines. 
The Traveling Salesman Problem (TSP)~\cite{dantzig1954solution}, a well-studied problem in operations research, is often used to model the planning challenges encountered by a single robot.
Later, this formulation was extended to include multiple vehicles (mTSP) in~\cite{bellmore1974transformation}. The mTSP is more suitable for large-scale applications but is more complex than the TSP. Several solutions have been proposed to solve this problem, such as the genetic algorithm (GA)~\cite{miloradovic2021gmp,glocal}, which considers tasks that require specific vehicle types, and~\cite{chen2022consensus}, which uses a consensus-based bundling algorithm for limited replanning, but few works have included communication restrictions and failures. An approach, presented in~\cite{otte2020auctions}, utilizes an auction allocation algorithm to assign tasks but assumes enough locally connected robots to perform the assigned tasks. Other works, such as~\cite{gao2022meeting}, address the issue of prolonged disconnections using rendezvous locations. However, this can lead to unnecessary communication and laborious backtracking. In these environments, robots may operate with outdated or incomplete information while also being aware of the possibility of misinformation. A robot may act on information that it believes to be accurate, only to discover later that it is outdated or incorrect. This can have significant consequences, particularly in critical applications such as disaster response or military operations. In~\cite{al2018generation}, system failures are considered in the multi-agent policy search, but it is assumed that robots can communicate these disruptions.

In contrast, this work applies dynamic epistemic logic (DEL)~\cite{van2007dynamic}, allowing each robot in the MRS to reason and plan using its beliefs of other robots in the system while disconnected, updating its beliefs and policy if new events are observed, and routing to communicate when necessary. DEL is a formal logic that describes how beliefs and knowledge change and has recently been integrated into robotics applications. The method presented in~\cite{bolander2021based} recreates the Sally-Anne psychological test for human-robot interactions. Typical DEL-based multi-agent research uses epistemic planning for game theory-based policies~\cite{maubert2021concurrent}. We evaluate the use of DEL for an MRS application, equipping each robot with the ability to reason about the system's state, considering local observations. Our proposed solution expands on previous mTSP research. It shows that intermittent rendezvous will allow the MRS to reason about the system's state and share any new knowledge, updating its beliefs at runtime utilizing an epistemic planning framework. 
\section{Preliminaries}\label{sec:prelims}
\label{sec:preliminaries}
\subsection{Notation, Communication, \& Control}
Consider an MRS of $m$ robot in the set $\mathcal{A}$. 
We assume that all initial positions of the robots are known. The MRS's communication range is denoted as $r_c$, and a robot $i$ and $j$ can communicate when within this range.

The variable $\mathcal{V}_i\subseteq\mathcal{V}$ represents the subset of all tasks $\mathcal{V}$ assigned to a robot $i$. We let $\bm{x}_i(t)$ denote the state variable of the robot $i$ that evolves according to general dynamics:
\begin{equation*}
   \bm{x}_i(t+1) = \mathbf{g}(\bm{x}_i(t),\bm{u}_i(t),\bm{\nu}_i(t))
   \label{eq:NLDynamics}
\end{equation*}
where $\bm{u}_i\in\mathbb{R}^{d_u}$ and the variable $\bm{\nu}_i\in\mathbb{R}^{d_\nu}$ denote the control input and zero-mean Gaussian process uncertainty at time $t$. The tuple $\Omega^i = (q_i,\mathcal{V}_i)$ is called the robot $i$'s \textit{disposition} and is defined by the capability of the robot $i$ and its assigned tasks. The capability $q_i$ of robot $i$ is informed by its kinematic specifications, such as maximum velocities and physical dimensions.

\subsection{Epistemic Logic}
\label{sec:epiLog}
For this application, the epistemic language, $\mathcal{L}(\Psi,\AP,\mathcal{A}$)
is obtained as follows in Backus-Naur form~\cite{knuth1964backus}:
\begin{equation}
    \phi \Coloneqq H(\eta) \ | \ \phi\land\phi \ | \ \neg\phi \ | \ K_i\phi \ | \ B_i\phi  \nonumber
\end{equation}
where $i\in\mathcal{A}$. $H\in\Psi$ with $\Psi$ being a set of functions that describe the system state. $\eta$ generally denotes function arguments. $\neg\phi$ and $\phi\land\phi$ are propositions that can be negated and form logical conjunctions, where $\phi\in \AP$ and $\AP$ is a finite set of atomic propositions. $K_i\phi$ and $B_i\phi$ are interpreted as ``robot $i$ knows $\phi$" and ``robot $i$ believes $\phi$", respectively.

The distributed knowledge and reasoning for robots in the system are modeled using epistemic logic~\cite{bolander2021based}. An epistemic state for $\AP$ is represented by the tuple $s = (W,(R_i)_{i\in\mathcal{A}}, L, W_d)$ where $W$ is a finite set of possible worlds, $R_i\subset W\times W$ is an accessibility relation for robot $i$ simplified to $R$ for reference to all robots, $L: W\rightarrow \AP$ assigns a labeling to each world defined by its true propositions, and $W_d\subseteq W$ is the set of designated worlds from which all worlds in $W$ are reachable.
The initial epistemic state is denoted as $s_0 = (W,R,V,\{w_0\})$. If $W_d=\{w_0\}$, $s_0$ is the global epistemic state. The world, $w$, signifies a set of true propositions that, in our application, is the disposition of each robot. The worlds in which the system is described by the combinations of all possible dispositions of each robot in the MRS (e.g., task assignment, velocity).
The truth of $\mathcal{L}$-formulas in epistemic states is defined with standard semantics similar to \cite{bolander2021based}:
\begin{align*}
    (W,R,L,W_d) \models \phi &\text{  iff  } \forall w\in W_d, (W,R,L,w) \models \phi\\
    (W,R,L,w)\models \phi &\text{  iff  } \phi\in L(w) \text{ where } \phi\in \AP \\
    (W,R,L,w)\models K_i\phi &\text{  iff  } \forall v\in W, \text{ if } (w,v)\in R_i \\
    &\text{ then } (W,R,L,v)\models \phi \\
    (W,R,L,w)\models C\phi &\text{  iff  } \forall v\in W, \text{ if } (w,v)\in \cup_{i\in\mathcal{A}} \\
    &\text{ then } (W,R,L,v) \models\phi
\end{align*}
The accessibility relation $R_i$ represents the uncertainty of robot $i$ at run-time for a global epistemic state $s=(W,R,L,w_d)$. In this state, the robot $i$ cannot distinguish between the actual world $w_d$ and any other world $v$ where $(w_d,v)\in R_i$. Consequently, robot $i$'s knowledge is based on what is true in all of these worlds $v$. Sequences of relations are used to represent higher-order knowledge. For example, the statement ``robot $i$ knows that robot $j$ knows $\phi$'' is true in $s$ if and only if $s\models K_i K_j\phi$. This condition is satisfied when $\phi$ is true in all worlds accessible from $w_d$ through the composite relation of $R_i$ and $R_j$. The perspective of robot $i$ is defined as $s_i=(W,R,L,\{v \ | \ (w,v)\in R_i; w\in W_d\}$. If $s$ is the global state, then $s_i$ is the perspective of robot $i$ on $s$. In this work, we represent a subset of these perspectives as particles moving through the environment in the set
\begin{equation}
    \mathcal{P}_i=\{p_{ij,b} \ \forall j\in\mathcal{A},\forall b\in\mathcal{B}\}.
    \label{eq:particles_set}
\end{equation}
where beliefs $b\in\mathcal{B}$ are a finite set of particles for each robot $i$ that represent how a robot $j$ would behave given a different, but accessible, world $w\in W_d$.

Dynamic epistemic logic is expanded from epistemic logic through action models~\cite{bolander2021based}. These models affect a robot's perception of an event and influence its set of reachable worlds, $R_i$. A robot may plan to reduce the run-time uncertainty by taking actions. 
We simplify the notation of the action model by referring to actions in plain language. The action library, $A$, is the set of actions that a robot can enact during mission execution.
We express the epistemic product model as $s\otimes i:a = (W',L_i',V',W_d')$ where $s\otimes i:a$ represents the new epistemic state after the action $a\in A$ has been enacted by robot $i$. A planning task is represented by the tuple $\Pi = (s,A,\gamma)$. An execution policy $\pi$ is a sequence of actions in $A$ for robots in the MRS that will satisfy the common mission objective $\gamma$ given an epistemic state $s$.
\section{Problem Formulation}\label{sec:prob}
\vspace{-2pt}


In this work, we assume that all robots know the location of all tasks $\mathcal{V}$ present in the environment and the initial location of all robots in the system. Given a limited communication range $r_c$, this approach aims to minimize the total time to complete all tasks in the environment since robots can experience failures or disturbances during operation. We formally define our problems as:
\begin{problem}\label{problem1}{\bf{\emph{Centralized mTSP Planner with Intermittent Communication:}}}
Design a strategy for an MRS to complete all tasks while weighing efficient rendezvous points. The goal is to minimize the mission's duration, considering that faults and disturbances may occur during execution, necessitating a communication and replanning mechanism.
\end{problem}

\begin{problem}\label{problem2}{\bf{\emph{Robust Online Replanning:}}} 
Formulate a policy for robust online replanning of the team's operations when faults or disturbances decrease the original plan's efficiency. The policy should minimize the time to complete all tasks, considering any necessary communications with disconnected robots and the deprecated state of the system.
\end{problem}



The mathematical formulation of Problem~\ref{problem1} matches the one of mTSP~\cite{necula2018balancing} where we seek to minimize the longest tour of any robot represented by $\mathcal{Q}$. However, in our work, robots can have different starting and ending depots. We also allow robots to have no tasks assigned to them. 
The optimization problem is expressed in its epigraph representation, where the objective function is included in the constraints as
\begin{align}
\min \ \ & \ \mathcal{Q}\label{eq:objconstr}\\
\text{s.t.} \ \ &\sum_{i \in  \mathcal{V}^\Sigma} \sum_{j \in \mathcal{V}^\Delta} \omega_{ijs} \cdot x_{ijs} \leq \mathcal{Q}, 
\qquad \forall s \in \mathcal{A},\nonumber
\end{align}
where $\omega_{ijs}$ is the cost of traveling from task $i$ to task $j$ for robot $s\in\mathcal{A}$. The binary decision variable $x_{ijs}$ defines if the robot $s$ travels from task $i$ to task $j$. The sets $\mathcal{V}^\Sigma$ and $\mathcal{V}^\Delta$ represent the inclusion of all the tasks and all the starting depots, and all the tasks and the ending depot, respectively.
%
%
%
%
The goal of the optimization process is to minimize the variable $\mathcal{Q}$, also known as ``minMax'' optimization, where we minimize all robots' maximum tour or makespan.

\section{Approach}\label{sec:approach}
\begin{figure}[ht!]
\vspace{-5pt}
    \includegraphics[width = 0.45\textwidth]{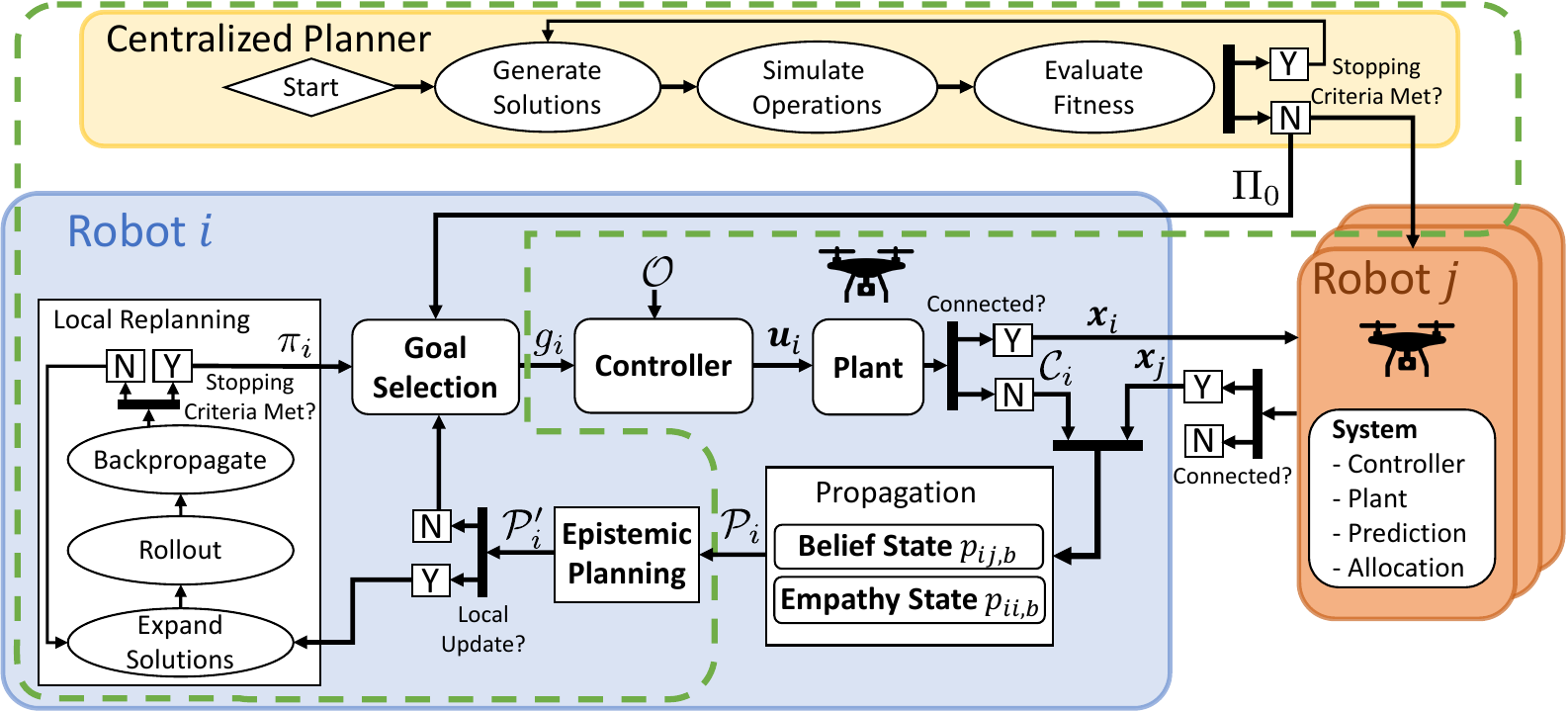}
    \vspace{-5pt}
    \caption{Diagram of the proposed approach. The contributions of this paper are within the green box.}
     \vspace{-10pt}
    \label{fig:mainFrame}
\end{figure}

Our proposed framework is designed for an open mTSP in which robots are not required to return to their starting location; instead, each robot has a starting and ending depot. When solving the mTSP, we promote intermittent communication by rewarding robot interactions during their respective tours, allowing robots to share information or realize that the original plan has changed. To realize changes, each robot propagates belief and empathy states to allow robots to observe the local environment, reason about system operations while disconnected, and adjust local plans when necessary. For ease of discussion, let us consider two robots, $i$ and $j$. From the perspective of the robot $i$, a {\em belief state}, $p_{ij,b}\in\mathcal{P}_i$, represents a possible state of a robot $j$ and an {\em empathy state}, $p_{ii,b}\in\mathcal{P}_i$, is robot $i$'s belief about itself from the perspective of other robots. With this knowledge, robot $i$ predicts and tracks empathy states to decrease the number of locations in which robot $j$ may search for robot $i$ and allows the system to complete all tasks more efficiently, given new operational constraints. The diagram in Fig.~\ref{fig:mainFrame} summarizes this architecture, where the centralized planner first routes robots to tasks in the environment, assessing the solution's fitness by minimizing the maximum tour length of a robot and rewarding intermittent communication based on a user's preferred settings. If robots disconnect, the set of belief and empathy particles, $\mathcal{P}_i$ propagates according to the sequence of actions, $\pi_{0}$ set by the centralized planner. If the robot locally observes system changes, epistemic planning allows each robot to determine the best series of actions to estimate the positions of lost robots, share any necessary information with other robots, and navigate to the remaining cities. 

 \subsection{The Centralized Planning Algorithm}\label{sec:centralizedPlan}

\begin{figure}[!t]
    \centering\includegraphics[width=0.44\textwidth]{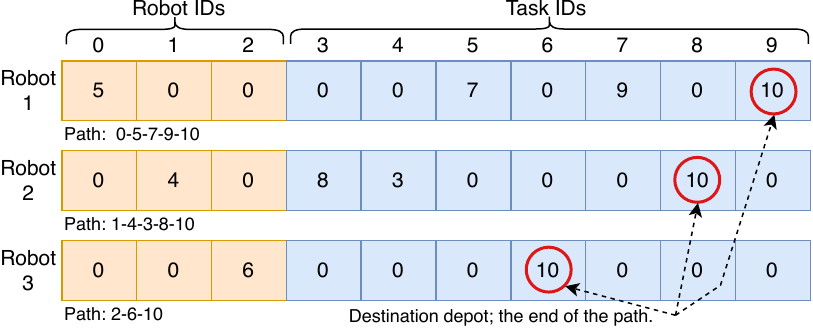}
    \vspace{-7pt}
    \caption{Graphical representation of chromosome encoding.}
    \label{fig:chromo}
    \vspace{-20pt}
\end{figure}

The centralized planner used in this work is based on a GA adapted to solve combinatorial optimization problems, specifically mTSP. 
%
Chromosomes are encoded as a set of arrays, where each array encodes a robot's plan. 
A graphical representation of a single chromosome is given in Fig.~\ref{fig:chromo}. 
The size of each array is equal to the sum of the number of robots and tasks, that is, $n + m$. 
The elements of the array are integer task IDs. 
Following the task chain, the robot's route can be extracted from the encoding. 
For example, in Fig.~\ref{fig:chromo}, if we look at Robot 1, we can see that the first task in its plan is 5, and the next task ID is then stored in column 5, which is Task 7. This continues until a destination depot with the ID of $n+m$, $10$ in this example, is reached. 
The initial population is generated randomly to start with a high diversity and seeded in the feasible region of the search space. 

The crossover operator is a modified version of Edge Recombination Crossover (ERX)~\cite{whitley1989scheduling}. 
The first step is to select two parents for crossover from the mating pool. 
The mating pool is generated, accounting for the crossover probability and each individual's fitness. Next, an adjacency matrix that contains the makeup of neighboring tasks based on the two chosen parents is constructed. 
We then randomly select a starting task and the selection chain continues by randomly selecting a task from a neighboring list of previously allocated tasks.
We randomly select a new task if all neighboring tasks are already allocated. 
We apply a jump mutation that randomly changes the placement of a single task in the route, and the swap mutation selects two tasks and swaps their locations.
Jump and swap mutations are invoked twice: the first for intra-robot mutations and the second for inter-robot mutations, such that there are both local and global mutation operations.

Greedy Search (GS) and 2-opt~\cite{junger1995traveling} are two local refinement methods implemented to reorder cities within a robot's plan resulting from the GA allocation of cities to salesmen. Local refinement methods exploit the candidate solution by reordering the list of tasks governed by the nearest-neighbor or 2-opt. In the next sections, we explain our modifications to increase the robustness of the MRS.


\subsection{Interaction Reward Mechanism}

The general rule for creating a good plan for TSP or mTSP is to have routes that do not cross. The most successful heuristic for solving these problems directly exploits this rule, e.g., 2-opt or Lin–Kernighan heuristics~\cite{lin1973effective}, but, in this work, we take a different approach by allowing the planner to create {\em interactions} between robots. Within this framework, we define an {\em interaction} as an event where robots are within the range $r_c$ to exchange information. 
Rewarding robots who travel within $r_c$ can create crossings in the resulting routes, contrary to~\cite{lin1973effective}. However, we argue that this can benefit the overall execution time of the mission when the system does not operate as planned. This will enable robots to detect system failures faster during execution without laborious backtracking after reaching the depot. 

To maximize the number of interactions among robots, we introduce a mechanism to reward the exchange of information between robots. However,
maximizing the number of interactions alone is not sufficient, as each interaction's value must be taken into account. For example, exchanging information at the beginning or close to the end of a mission may not be beneficial, as little new information can be gained from these interactions. Furthermore, redundant interactions over small-time intervals should not be highly rewarded since no new information is likely to be shared. To capture this, we introduce, for every robot $i$, a time interval $[\tau_{i}^{s},\tau_{i}^{e}]$ when the robot can be rewarded for interacting with other robots.
The potential reward is designed to grow linearly from $t=\tau_{i}^{s}$ for $\Phi_{i} = (\tau_{i}^{e} - \tau_{i}^{s})/2$ time units and to stay constant for the remaining part of the interval. Then, the potential reward function $U_i(x)$ for robot $i$ is defined as:
\begin{align}
U_{i}(x) = 
\begin{cases}
x, & \textmd{if} \;  0 < x \leq \Phi_{i}  \\
\Phi_{i}, & \textmd{if}\; \Phi_{i} < x \leq 2 \Phi_{i},\\
0,  & \textmd{otherwise.}
\end{cases}
\label{eq:potential_reward}
\end{align}
However, the actual reward $\mathcal{R}_i(t)$ is assigned only if the interaction happens, according to
\begin{align}
&\mathcal{R}_i(t) =
    \begin{cases}
    \rho U_{i}(t-t_{i}^{lr}), & \text{if robot $i$ interacts at time $t \in [\tau_{i}^{s},\tau_{i}^{e}]$},\\
    0, & \text{otherwise},
    \end{cases}
    \label{eq:reward}
\end{align}
where $t$ is the current time, and $t_{i}^{lr}$ is the time when the last reward was assigned to robot $i$. 
Furthermore, $\rho$ is half of the average distance between tasks, and it is introduced as a weight related to the structure of the problem. This means that $\rho$ scales with the problem instance.
The total reward is then calculated as:
 \begin{equation}
\mathcal{R}_{\textit{tot}} = \sum_{t=0}^{\mathcal{Q}}\sum_{i \in \mathcal{A}} \mathcal{R}_i(t).\label{eq:total_interaction_reward}
 \end{equation}
However, two robots may exchange information frequently while a third robot is in no contact with them. To overcome this issue, we also introduce a penalty mechanism for robots not interacting with other robots. This mechanism requires a tunable threshold, $\sigma$, to be defined, e.g., all robots have to interact with another robot at least once before completing 50\% of a given mission, i.e., $\sigma = 0.5$. The penalty for failing to do so is calculated as follows:
\begin{align}
P_{i} = 
\begin{cases}
(t_{i}^{\textit{int}} - \sigma \cdot t_{i}^{\textit{max}}) \cdot \rho & \textmd{if}\; t_{i}^{\textit{int}} > \sigma \cdot t_{i}^{\textit{max}},\\ 
0,  & \textmd{otherwise,}
\end{cases}
\label{eq:penalty}
\end{align}
where $t_{i}^{\textit{max}}$ is the time required for robot $i$ to complete its plan, and $t_{i}^{\textit{int}}$ is the time when robot $i$ had its first interaction with another robot. 
To get the total penalty, $P_{\textit{tot}}$, we sum up the penalties over all robots.
The optimization problem \eqref{eq:objconstr} can now be updated with rewards and penalties as follows:
\begin{align}
&{\min} \ \ \mathcal{Q} - \mathcal{R}_{\textit{tot}} + P_{\textit{tot}}.\label{eq:optFinal}
\end{align}
We also extend the aforementioned approach if a user requires more control over the mission makespan compared to a traditional mTSP solution. In this case, we solve a bilevel optimization problem where we first minimize $\mathcal{Q}$ subject to~\eqref{eq:objconstr} and then optimize the following:
\begin{align}
    \max \ \ &\mathcal{R}_{tot} - P_{tot} - \Delta\mathcal{Q}    \label{eq:userDefinedOpt}\\
    \text{s.t} \ \ &\eqref{eq:objconstr}; \ \Delta\mathcal{Q}\leq\delta\mathcal{Q}^* \nonumber
\end{align}
where $\Delta\mathcal{Q}$ represents the difference between the solution to the upper-level optimization problem~\eqref{eq:objconstr} represented by $\mathcal{Q}^*$ and the inner optimization task in~\eqref{eq:userDefinedOpt}. The user-defined variable $\delta\in[0,1]$ represents the extent to which the typical mTSP makespan minimization can be worsened to increase interactions using~\eqref{eq:total_interaction_reward} and~\eqref{eq:penalty}. 
In this way, we have better control over the quality of the produced solution, with the mission duration upper-bounded by the user-defined limit. 

\subsection{Belief \& Empathy Propagation}
 \label{sec:bepropagation}
So far, we have explained how we developed a centralized strategy that allows intermittent interactions. Now, we transition to online adaptations, indicated by the blue section in Figure~\ref{fig:mainFrame}, to plan based on the information gained from these interactions. 
In our framework, each robot propagates belief states for all robots in the MRS. This allows a robot $i$ to plan according to its beliefs about other robots and to empathize with what other robots expect robot $i$ to do while disconnected. Each robot predicts the future states of a set of beliefs for all robots in the system and will follow the closest empathy state even if a malfunction occurs, allowing a robot only to propagate a finite number of beliefs represented by the set $\mathcal{P}_i$ in~\eqref{eq:particles_set}. A robot $i$ defines its empathy particles as $\mathcal{P}^e_{i} = \{p_{ii,b} \ \forall b\in \mathcal{B}\}$ and its belief particles about other robots as $\mathcal{P}^r_{i} = \{p_{ij,b}  \ \forall j\in \mathcal{A},\forall b\in \mathcal{B}\}$. 
If disconnected, a robot $i$ propagates beliefs according to the last globally communicated epistemic state between robot $i$ and robot $j$, moving particles based on how robots would behave given a subset of true propositions from the set $\AP$ introduced in Sec.~\ref{sec:epiLog}. 
Initially, we note that all robots know the initial position and disposition of all robots, defined by the centralized plan in Sec.~\ref{sec:centralizedPlan}. Particles are propagated along the tour provided by the centralized planning algorithm. Given that all robots follow an empathy particle during exploration, we next present our strategy to update the epistemic state if changes occur at runtime.

\subsection{Epistemic Replanning}
Robots follow the centralized plan initially, but if operations do not go as planned and a robot experiences a failure or other robots communicate changes to the system, a rational belief update must occur. We formulate a belief update for this application as:
(i) A robot updates its own belief given it cannot operate as expected;
(ii) a robot updates its belief about another robot, given it is not traveling according to a previously expected belief;
(iii) a robot communicates a belief update about a disconnected robot to a subset of connected robots within the communication range.
In all these scenarios, a robot can update its execution policy of tasks in the environment, given its new belief about the MRS to complete all tasks in the environment, which is equivalent to satisfying the common goal $\gamma$ from Sec.~\ref{sec:epiLog}. However, belief updates can have cascading effects across the MRS if new information is not communicated efficiently and on time. Therefore, we introduce a hierarchical framework to update allocations at runtime and when failures or disturbances occur and robots can no longer follow the original plan.

\subsubsection{Epistemic Updates}\label{sec:epiUpdates}
Establishing a mechanism for logical updates is important to determine when or if a robot should find or communicate with other members of its team and how to reach a consensus on disconnected team members within a locally connected team. A dynamic epistemic logic (DEL) framework allows a robot $i$ to succinctly share beliefs and update a robot's perspective $s_i$ on the epistemic state $s$. There are two cases where updates can occur: i) when connected to all robots and ii) when expecting to connect with another robot. From the established semantics in Sec.~\ref{sec:epiLog}, we know $A=\{percieve(\phi), announce(\phi),complete(\phi)\}$ 
The action {\em complete} represents a robot completing a task, {\em perceive} symbolizes a robot observing a generic proposition $\phi$ about the MRS, and {\em announce} constitutes communication with a locally connected team.  The set $\Psi=\{track\}$ is functionally interpreted for $B_i \, track(p_{ij,b})$ as the robot $i$ knows that the robot $j$ is tracking the belief particle $b$.

First, we address a belief update when robots are within communication range. We assume that because robots are cooperative, all belief updates are accepted and are only outdated if an event occurs, such as system failures or disturbances. These updates are announced such that the epistemic state from robot $i$'s perspective is:
\begin{equation}
    s_i\otimes announce(\Omega) = s'_i\models K_i V(\Omega^i) \bigwedge_{j\in\mathcal{A}} K_i K_j V(\Omega^j) \ \forall i\in\mathcal{C}.
    \label{eq:publicAnnounce}
\end{equation}
where $announce(\Omega)$ is an action symbolizing the announcement of all robots' dispositions, $\Omega$. 
The notation models robot $i$'s knowledge of the dispositions of all robots, and the function $V(\Omega^i)$ maps the dispositions of the robot $i$ to atomic propositions in the set $\AP$. The set $\mathcal{C}\subseteq\mathcal{A}$ represents the set of robots within robot $i$'s communication range.
The belief particles are updated from the announcement of all states to the MRS such that
\begin{equation}
    p_{ij,b}\gets \Omega^j, \ \forall (i,j)\in \mathcal{A}^2, \ \forall b\in \mathcal{B}.
\end{equation}

Since beliefs are shared according to \eqref{eq:publicAnnounce}, the particles in this set are propagated according to the dispositions of each robot. For example, in a three-robot team, if robot 1 communicates with robots 2 and 3 that it will execute robot 3's tasks, all robots would propagate a belief particle that moves robot 1 according to its assigned tasks.

The {\em perceive} action causes a robot to change his belief in the epistemic world. If robot $i$ perceives that robot $j$ is not at its believed location, it updates its epistemic state with the epistemic action {\em percieve}:
\begin{equation}
s_i\otimes i:perceive(\neg track(p_{ij,b}))\models V(\Omega^j,B_i \neg track(p_{ij,b}))
\end{equation}
where the function $V$ takes two arguments, mapping robot $i$'s updated belief about robot $j$ to an atomic proposition in $\AP$ and robot $i$'s new epistemic state is evaluated as the epistemic product after {\em perceive} has been enacted. The particle propagation does not change in this case since robots may seek out other robots without knowledge of this belief update. In the event of a malfunction or fault of a robot, the robot updates its belief in the same way with $j = i$ and tracks respective empathy particle $p_{ii,b+1}$.

In this way, the knowledge of disconnected robots is not affected, nor does the robot $i$ update its belief that a disconnected robot would know the updated information. With our epistemic states and actions defined, we now describe how these concepts can be used for planning. As stated in Sec.~\ref{sec:epiLog}, a planning task for the MRS is defined by the tuple $\Pi=(s,A,\gamma)$ where $\gamma$ is a goal formula. In plain language, the goal formula is to complete all tasks. We define the epistemic update associated with a robot who completes an assigned task $\nu\in\mathcal{V}$ as:
\begin{equation}
    s_i\otimes i:complete(\nu)\models V(\Omega^i,B_j complete(\nu)) \ \forall j\in\mathcal{A}.
\end{equation}
The variable $\Omega^i$ is also updated to represent the new disposition of robot $i$. 
Given that other robots are also tracking the believed location of robot $i$, belief updates occur without communication, although these beliefs may be incorrect if a failure has occurred. Thus, an augmented policy that allows the MRS to achieve the common goal must be enacted.

\begin{figure}[ht!]
    \centering
    \subfigure[Planning for a dynamic task]{
    \fbox{\includegraphics[width=0.22\textwidth,trim={9cm 7cm 9cm 6.7cm},clip]{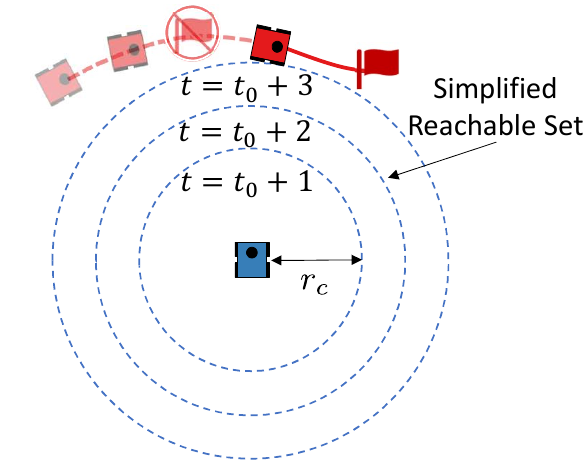}}
    \label{fig:commAliveBot}
    }\hspace{-0.52em}%
    \subfigure[Finding at an unknown location]{
    \fbox{\includegraphics[width=0.22\textwidth,trim={9cm 7cm 9cm 6.7cm},clip]{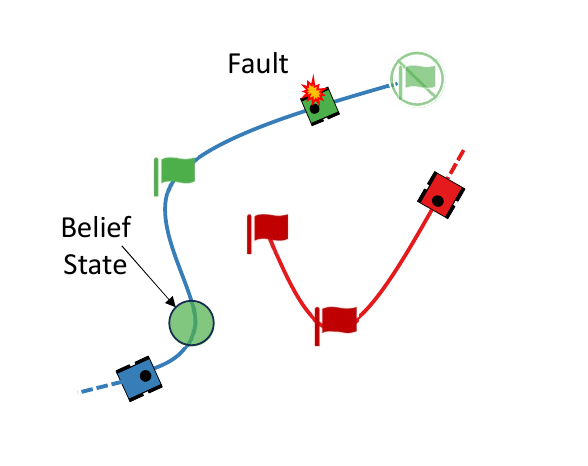}}
    \label{fig:commDeadBot}
    }%
    \vspace{-5pt}
    \caption{Examples of tasks generated as a result of belief updates}
    \label{fig:deadAndAliveFinds}
    \vspace{-5pt}
\end{figure}

In the event of a malfunction, we introduce two new types of tasks that allow operational robots to gather the necessary information about the system's condition and complete any unfinished tasks. These types of tasks are called {\em gossiping} and {\em finding}. Given robot $i$'s belief and the planned interactions with other robots according to the mTSP solution \eqref{eq:optFinal}, a robot should communicate before any planned interaction. Ensuring the completion of communication tasks (gossiping) and promptly identifying malfunctioning robots are vital steps to facilitate accurate information exchange and prevent the spread of misinformation within the system. The estimation of the interaction point can be determined using the time-based trajectory of the belief state and the reachable set of the robot involved, as depicted in Fig.~\ref{fig:commAliveBot}. The position of robot $j$'s belief state at a specific time $t$ is denoted as $p_{ij,b}(t)$. To find the point where robot $i$'s communication range intersects with the communication range of robot $j$'s predicted location, we find the minimum timestep $t_r$ that satisfies the equation:
\begin{equation}
    \|\bm{x}_i(t_r)-\bm{p}_{ij,b}(t_r)\| - R(t_r) > r_c
\end{equation}
where $\|\bm{x}_i(t_r) - \bm{p}_{ij,b}(t_r)\|$ represents the distance between the location components of the robot's position, $\bm{x}_i$ and robot $i$'s belief about robot $j$, $\bm{p}_{ij,b}$. The reachable set, $R$, for robot $i$ expands at every timestep based on the robot's velocity. If all belief states have been checked for a deprecated robot $j$, a robot $i$ backtracks along the previously established path until robot $j$ is located. An example is shown in Fig.~\ref{fig:commDeadBot}, where a blue robot routes backward along the green robot's path to communicate and reallocate any remaining tasks.

\vspace{-12pt}

\subsubsection{Balanced Workload Partitioning}
Given the limited nature of communication in this application, robots first assign new tasks to connected robots before optimizing their path~\cite{mazdin2021distributed} so that robots do not need to maintain a connection while optimizing routes to new tasks. Robots instead partition tasks based on a balanced workload and accounting for any belief updates. For example, suppose two robots, $i$ and $j$, are connected, and robot $k$ is not within the communication range. In that case, robot $i$ might believe that robot $k$ is functioning according to the initial state, $s_0$, but robot $j$ did not {\em perceive} robot $k$ at its respective belief state $p_{ik,b}$. So robot $j$ {\em announces} its belief to robot $i$. Robots $i$ and $j$ then bid on the new task, which is to find robot $k$. We let $\mathcal{V}_c$ be the set of tasks the connected robots must complete, and the cost function for allocating a task to a robot is user-defined (e.g., distance, time). Algorithm~\ref{alg:initPop} presents the bidding mechanism used in this paper, noting that this is only instigated if a belief update about the MRS functionality has occurred (i.e., a fault or disturbance). Once the task allocations have been determined, the next step is to find the optimal tour. 
\begin{algorithm}
\caption{Balanced Workload Partitioning}\label{alg:initPop}
\begin{algorithmic}[1]
\State $tour_s \equiv \emptyset \ \ \forall s\in \mathcal{C}$
\ForEach{$\nu\in\mathcal{V}_c$}
\ForEach{$s\in\mathcal{C}$}
\State $bid_s = cost(tour_s\cup \nu)$
\EndFor
\State $winner = \underset{s\in\mathcal{C}}{\argmin}(bid_s)$
\State $tour_{winner}\gets tour_{winner}\oplus \nu$
\EndFor
\end{algorithmic}
\end{algorithm}

\subsubsection{Monte Carlo Tree Search}
As mentioned in Sec.~\ref{sec:intro}, we combine Monte Carlo Tree Search (MCTS) with a DEL to implicitly coordinate plans when the system does not operate as originally intended. Referring to Sec.~\ref{sec:bepropagation}, to limit the policy search space, each robot's state consists of believing that a robot $j$ is following one of the particles represented in robot $i$'s set of particles $\mathcal{P}_i$.  MCTS is applied to complex games such as chess or Go to find the next best move, even in real-time~\cite{kocsis2006bandit}. To model the solution space effectively, we use the current epistemic state from a robot $i$'s perspective $s_i$. The MCTS algorithm simulates changes to the epistemic state when an action is taken and is represented as $s'_i\sim s_i\otimes a$. Robots add {\em gossiping} or {\em finding} tasks based on local observations, as discussed in Sec.~\ref{sec:epiUpdates} when 1) a robot experiences a fault or disturbance, or 2) a robot $i$ observes that its epistemic belief about the state of robot $j$ is incorrect. 

In this work, the search tree is generated by repeating the four steps -- {\em selection}, {\em expansion}, {\em simulation}, and {\em backpropagation} 
-- until a certain termination condition is met, which in this approach is a certain number of simulations. In the selection stage, a leaf node that has not yet been fully expanded is selected. We employ the upper confidence bound applied to trees (UCT) technique, which is typical in MCTS, to decide which vertex to simulate from the root node. 
Specifically, UTC was chosen because it has been shown to strike a good balance between exploration and exploitation~\cite{kocsis2006bandit}. Expansion occurs by randomly applying a random action or, in this case, adding a random task to a robot's route. The simulation then performs a random remaining route until termination (i.e., all of the robots' allocated tasks are performed) and then backpropagates the reward, applying the estimated value to the expanded node in the expansion step. The MCTS seeks to maximize the negative time it takes for a robot to complete all its assigned tasks, estimating the time to {\em find} and {\em gossip} with robots using the methods in Fig.~\ref{fig:deadAndAliveFinds}. We summarize the MCTS simulation applied in this approach in Algorithm~\ref{alg:MCTS}.
\begin{algorithm}[h]
\caption{MCTS - Simulate}\label{alg:MCTS}
\begin{algorithmic}[1]
\State $tour = \text{child.tour}(child\_rollout)$
\State $cost = 0$
\ForEach{$c\in tour$}
\If{$type(c) = $ "Static Robot"}
\State $path = $reverse($particle(c).tasks$)
\ElsIf{$type(c) = $ "Dynamic Robot"}
\State $path =$ find\_intersect\_point($particle(c).tasks$)
\Else
\State $path = $ task\_location($c$)
\EndIf
\State $cost \pluseq $ time\_to\_traverse($path$)
\EndFor
\State $reward = -cost$
\end{algorithmic}
\end{algorithm}
The tour with the lowest estimated cost is the chosen execution policy, $\pi$ for all the robots in the system that will satisfy $\gamma$.

\begin{figure}[ht!]
\vspace{-10pt}
    \centering
    \hspace{0.5cm}\subfigure{
    \includegraphics[width=0.43\textwidth,trim={5.8cm 18.2cm 3.5cm 9.2cm},clip]{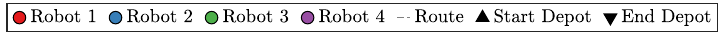}
    }\vspace{-0pt}
    \setcounter{subfigure}{0}
    \subfigure[\hspace{0.5em}Ideal: 61m makespan]{
    \includegraphics[height=3.1cm,width=3.3cm,trim={6.25cm 10cm 9.4cm 11.05cm},clip]{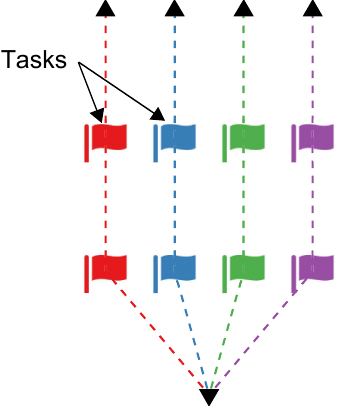}
    \label{fig:toyExampleIdeal}
    }%
    \hspace{1em}\subfigure[\hspace{-0.0cm} Ours: 69m makespan]{
    \includegraphics[height=3.1cm,width=2.7cm,trim={7.4cm 10cm 9.4cm 11.05cm},clip]{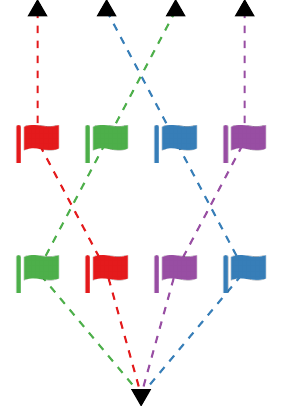}
    \label{fig:toyExampleEP}
    }
    \vspace{-7pt}
    \caption{Ideal mTSP allocation for 4 robots is shown in (a) and (b) is the solution with our proposed method.}
    \label{fig:toyExample}
    \vspace{-10pt}
\end{figure}

\begin{figure}[ht!]
    \centering
    \subfigure{
    \includegraphics[width=0.48\textwidth,trim={4.4cm 18cm 3cm 9.2cm},clip]{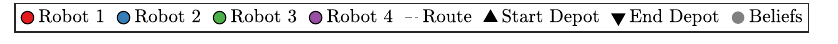}
    }\vspace{-10pt}
    \setcounter{subfigure}{0}
    \subfigure[]{
        \fbox{\includegraphics[width=0.14\textwidth,trim={8.7cm 9.8cm 7.8cm 11.5cm},clip]{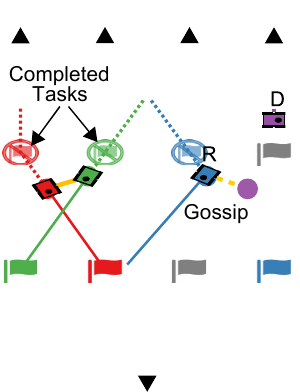}}\label{fig:toy_frame1}
    }\hspace{-0.55em}%
    \subfigure[]{
    \fbox{\includegraphics[width=0.14\textwidth,trim={8.7cm 9.8cm 7.8cm 11.5cm},clip]{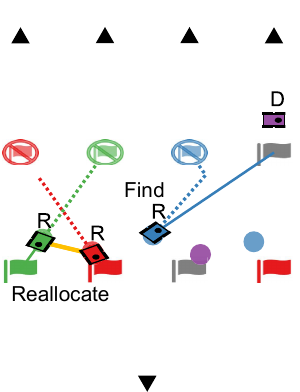}}\label{fig:toy_frame2}
    }\hspace{-0.55em}%
    \subfigure[]{
    \fbox{\includegraphics[width=0.14\textwidth,trim={8.7cm 9.8cm 7.8cm 11.5cm},clip]{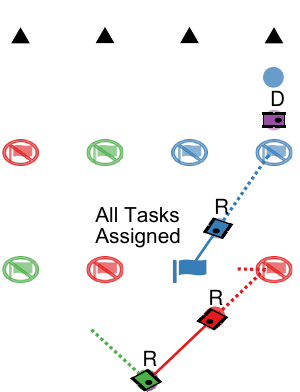}}\label{fig:toy_frame3}
    }%
    \vspace{-7pt}
    \caption{Our approach on a toy example, where the purple robot fails, and the blue robot realizes that the purple robot is not where expected.}
    \label{fig:toyExample_stages}
    \vspace{-10pt}
\end{figure}

To aid the reader in understanding, the proposed approach is implemented on the toy example shown in Fig.~\ref{fig:toyExample}.
We show the trivial solution to the mTSP in Fig.~\ref{fig:toyExampleIdeal} and the modified mTSP solution considering the interaction reward from \eqref{eq:total_interaction_reward} in Fig.~\ref{fig:toyExampleEP}. In Fig.~\ref{fig:toyExample_stages}, we show a subset of frames from the approach in which the blue robot realizes that the purple robot is deprecated in Fig.~\ref{fig:toy_frame1}, gossips the information to the red and green robots who reallocate remaining tasks while the blue robot is charged with finding the purple robot in Fig.~\ref{fig:toy_frame2}. In Fig.~\ref{fig:toy_frame3}, the blue robot has communicated with the purple robot and routes to the remaining task in the environment before returning to the home base. 
    \vspace{-5pt}

\section{Simulations}\label{sec:sims}
This section showcases the outcomes obtained through MATLAB simulations of our method executed by teams of two to five robots. The square environment used for the simulations has dimensions of $30\times 30, 30\times 30, 90\times 90,$ and $150\times 150$ [m], and for each scenario, a total of 10, 10, 30, and 50 tasks were generated. The locations of the tasks were randomly generated for each scenario.

\begin{figure}[ht!]
    \vspace{-15pt}
    \centering
    \subfigure[Scenarios with 1 failure]{
    \includegraphics[width=0.25\textwidth,height = 3.5cm,trim={4.5cm 9.32cm 6cm 9.43cm},clip]{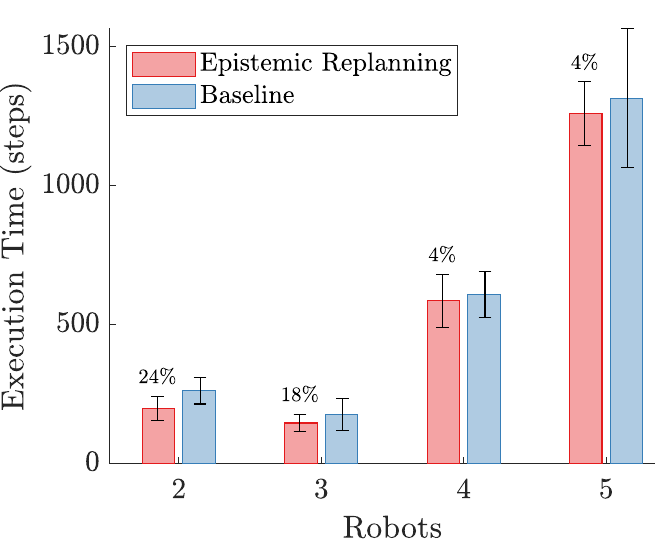}
    \label{fig:comparison1fail}
    }%
    \subfigure[Scenarios with 2 failures]{
    \includegraphics[width=0.24\textwidth,height = 3.5cm,trim={6.7cm 9.32cm 5cm 9cm},clip]{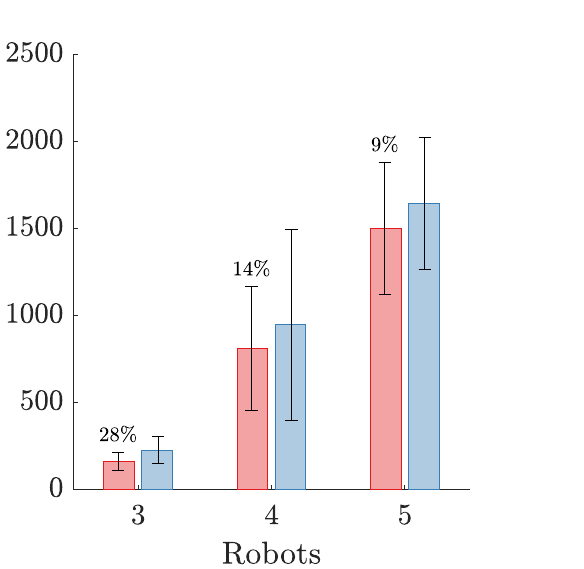}
    \label{fig:comparison2fails}
    }
    \vspace{-10pt}
    \caption{Comparison of a baseline heuristic and the proposed approach.}
    \label{fig:comparison}
    \vspace{-5pt}
\end{figure}

In our proposed approach, each robot propagates three particles. The initial maximum speed of each vehicle is 5 [m/s], and the second and third particles travel at a linear speed that is 80\% and 60\% of the vehicle's maximum speed, respectively. The maximum communication range is 5 [m] from the center of the robot. In our simulations, we implemented one fault for teams of two to five robots and two for teams of three to five robots, randomly occurring to any robot, causing the affected robot to track its second or third empathy particle or fail (i.e., zero velocity). Our approach was compared to a baseline heuristic in which the routes are determined by minimizing their makespan from~\eqref{eq:objconstr} and backtracking to find team members who do not arrive at the depot when expected. We let $\delta$ equal 30\% to increase the number of interactions between robots from~\eqref{eq:userDefinedOpt} such that the makespan of our solution can be up to 30\% longer than the baseline heuristic solution. We compare with this baseline to determine whether increased interactions and epistemic replanning truly improved the outcome.
As shown in Fig.~\ref{fig:comparison}, our approach outperforms the baseline heuristic by a significant margin in all scenarios, and we note that the margin increases as the number of failures increases between Fig.~\ref{fig:comparison1fail} and Fig.~\ref{fig:comparison2fails}.

\noindent \textbf{\textit{Discussion}} -- We emphasize that the margin of improvement is smaller as the teams become larger because information sharing becomes more inefficient as interactions between all robots become sparse. This introduces an interesting expansion outside the scope of this work for introducing optimal sub-teaming to create more efficient information sharing for static or dynamic teams during operations. In addition, not all vehicles are equally likely to fail. As vehicles age, they may become less reliable, requiring more dependable vehicles to take over or pick up additional tasks if a vehicle's operating capacity is deprecated during operations~\cite{stancliff2009planning}.

\begin{figure*}[ht!]
\centering
    \subfigure[]{
    \includegraphics[width = 0.155\textwidth,height=4.2cm,trim = {0cm 0.02cm 0cm 0},clip]{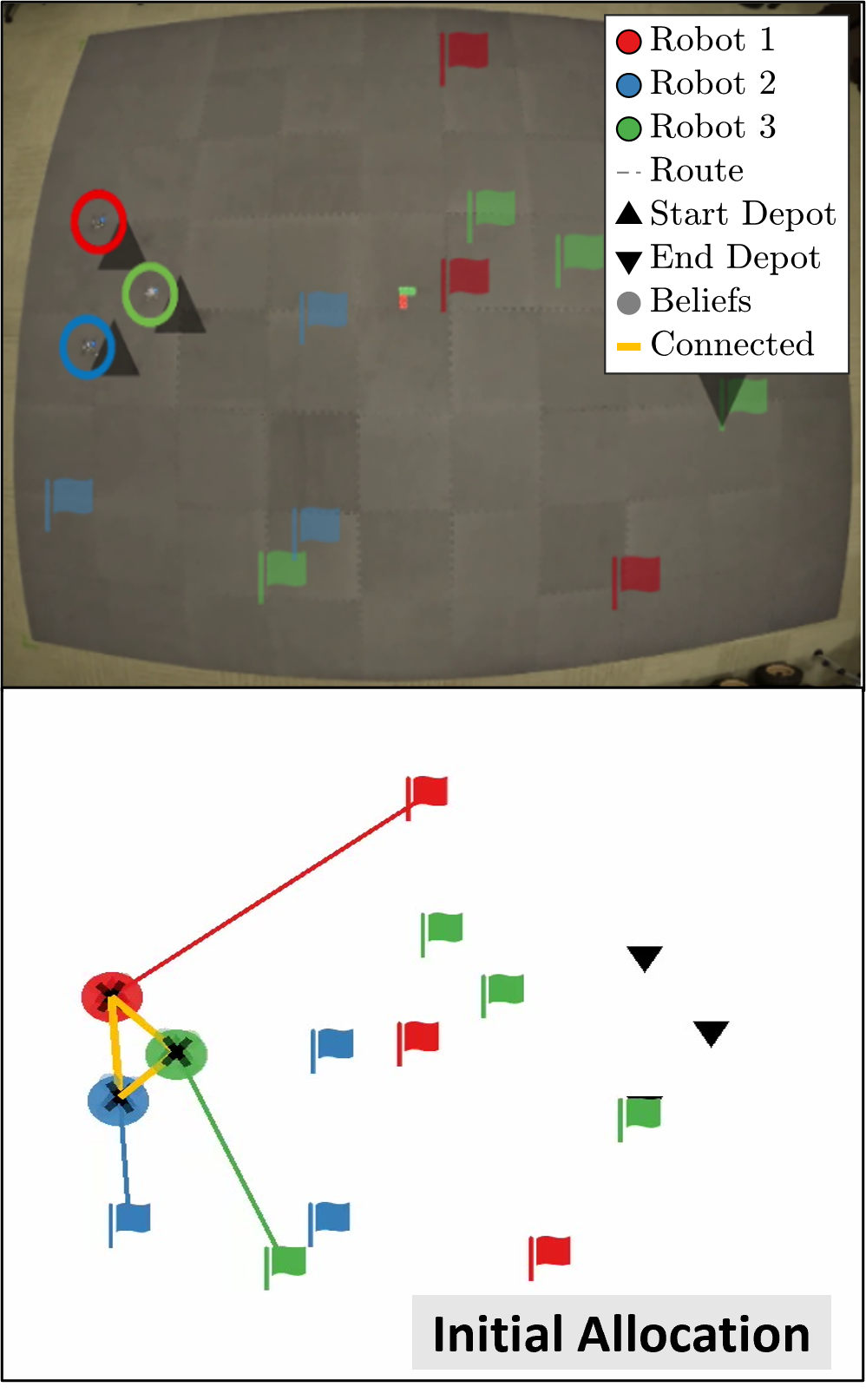}
    \label{fig:exa}
    }%
    \subfigure[]{
    \includegraphics[width = 0.155\textwidth,height=4.2cm,trim = {0cm 0.04cm 0cm 0},clip]{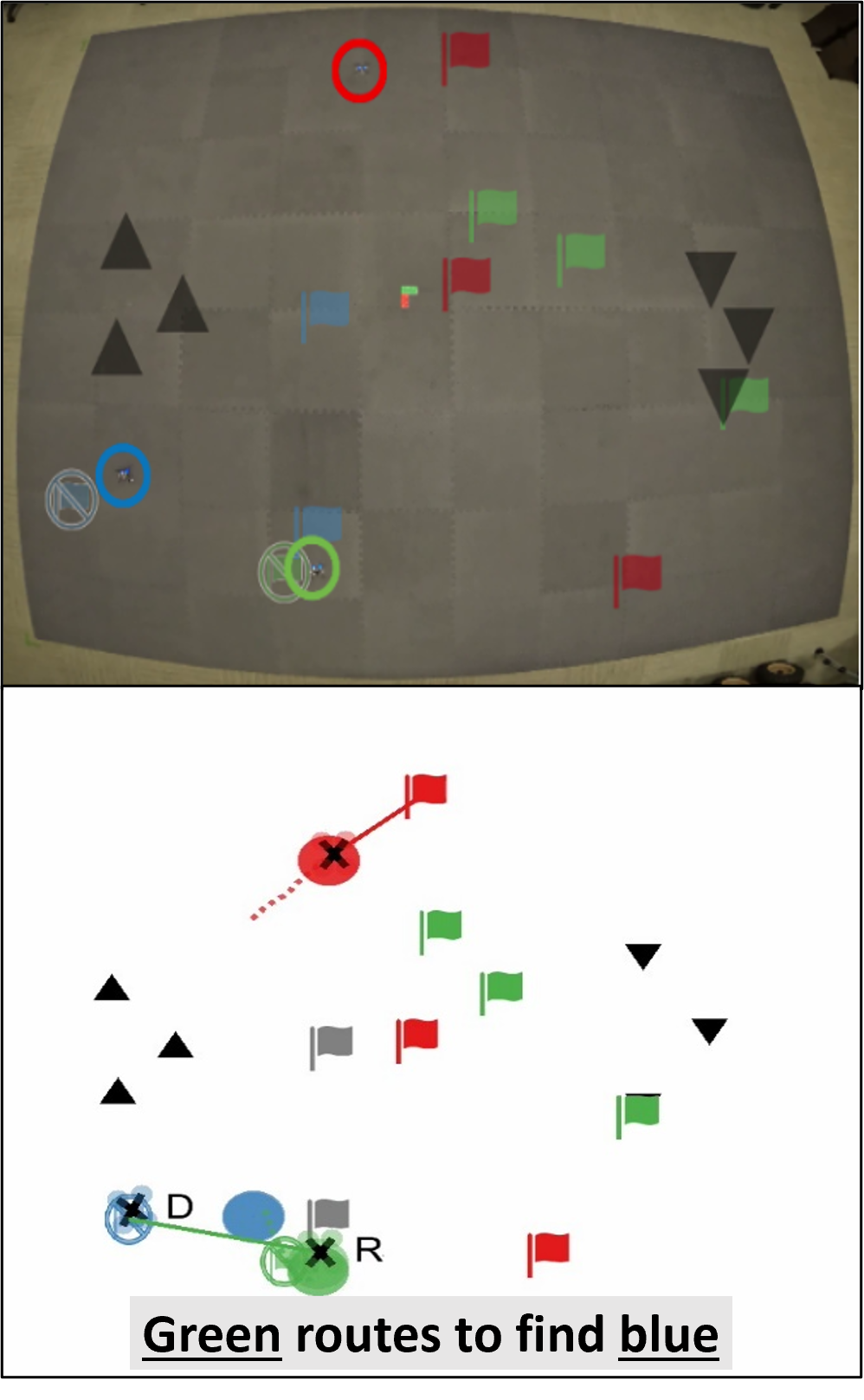}
    \label{fig:exb}
    }%
    \subfigure[]{
    \includegraphics[width = 0.155\textwidth,height=4.2cm,trim = {0cm 0.0cm 0cm 0},clip]{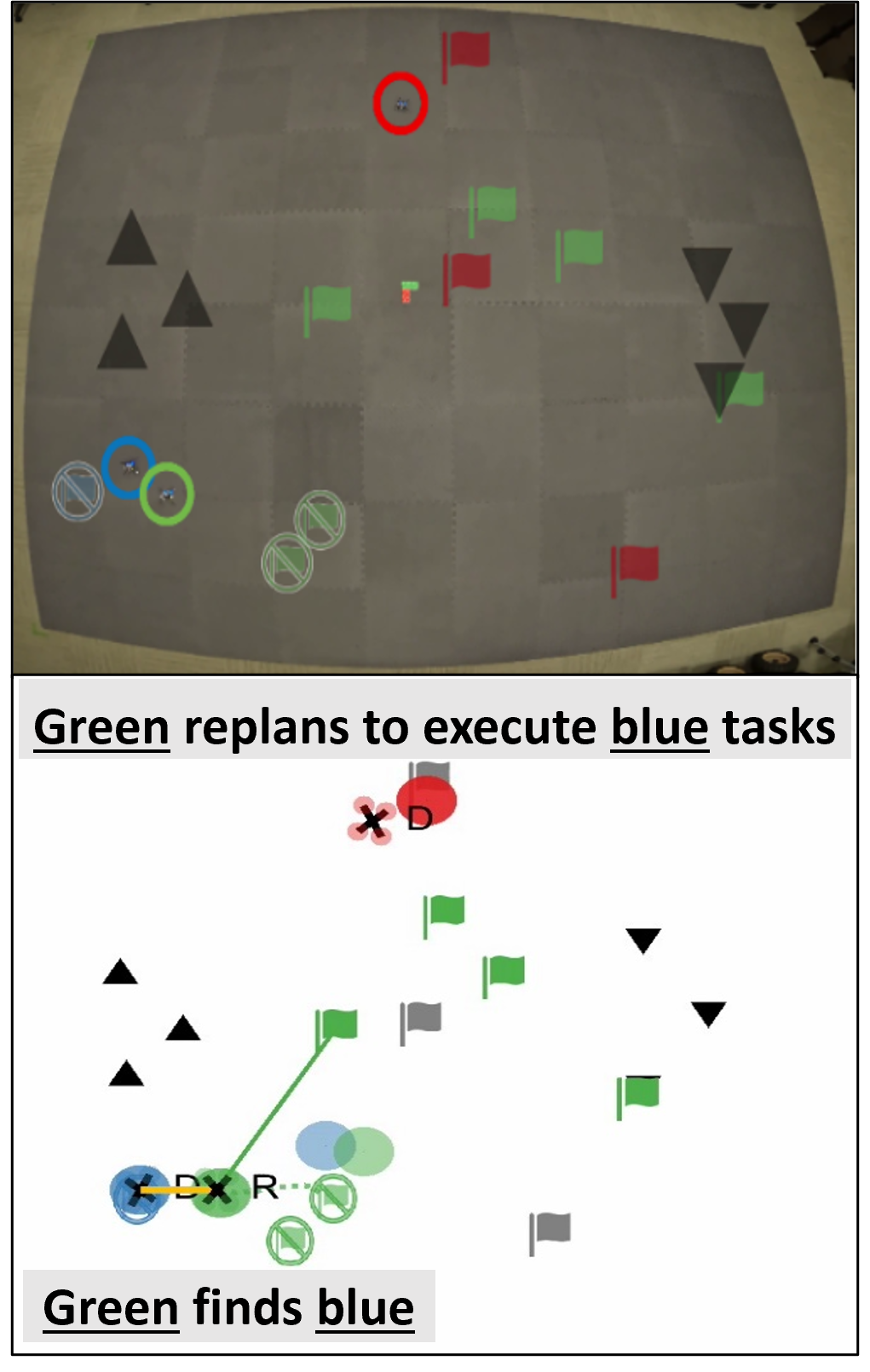}
    \label{fig:exc}
    }%
    \subfigure[]{
    \includegraphics[width = 0.155\textwidth,height=4.2cm,trim = {0cm 0.0cm 0cm 0},clip]{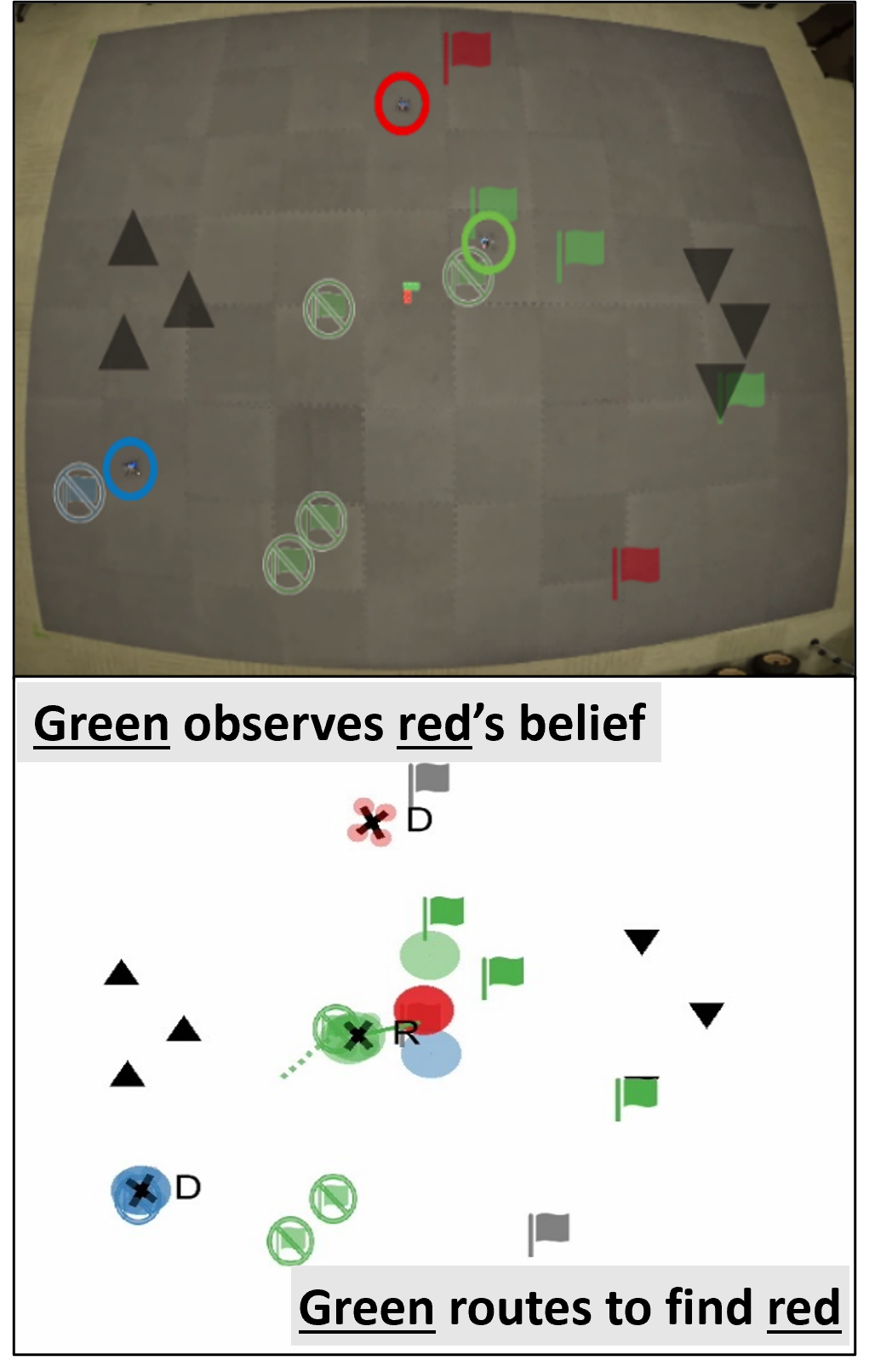}
    \label{fig:exd}
    }%
    \subfigure[]{
    \includegraphics[width = 0.155\textwidth,height=4.2cm,trim = {0cm 0.0cm 0cm 0},clip]{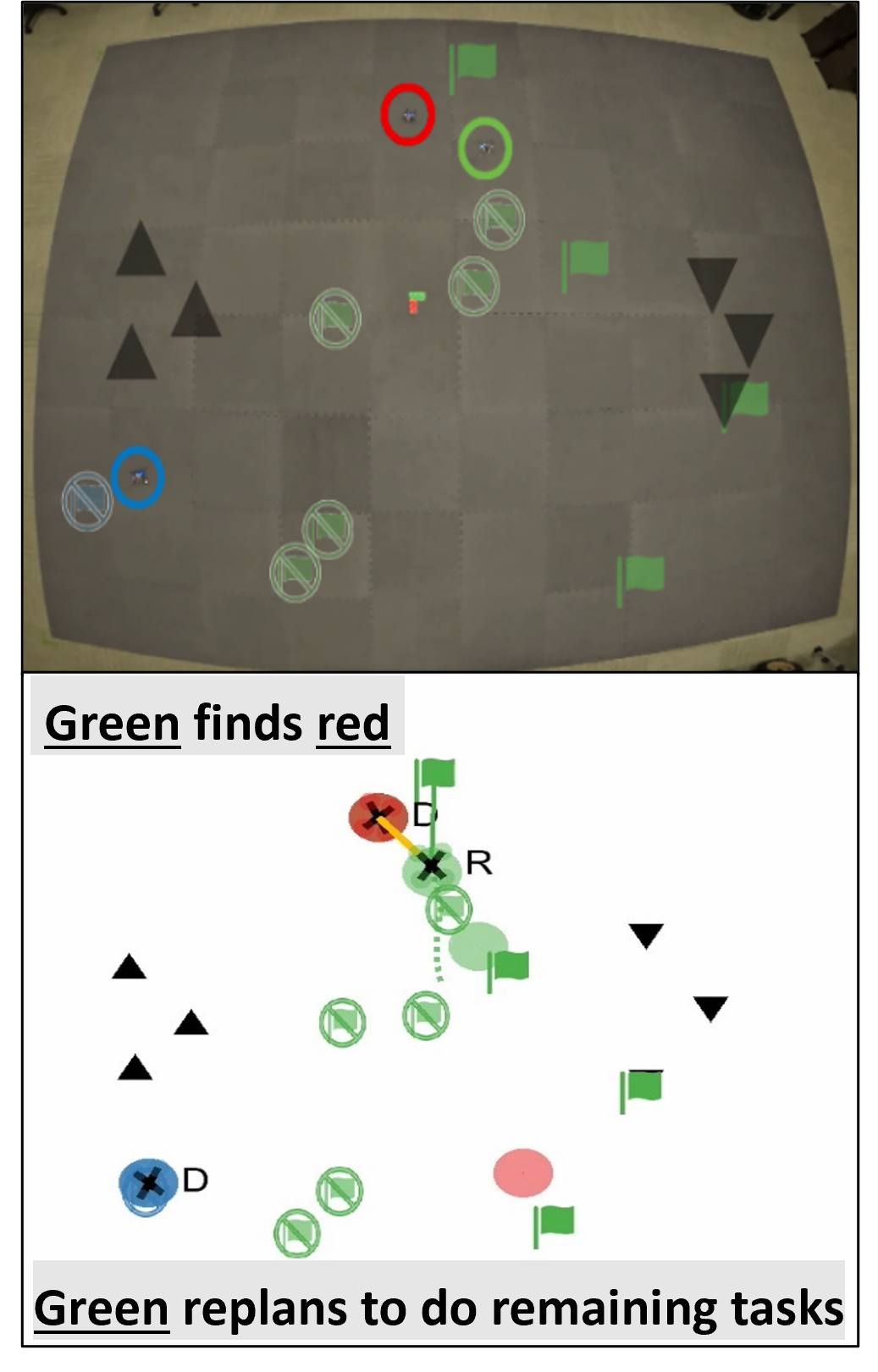}
    \label{fig:exe}
    }%
    \subfigure[]{
    \includegraphics[width = 0.155\textwidth,height=4.2cm,trim = {0cm 0.03cm 0cm 0},clip]{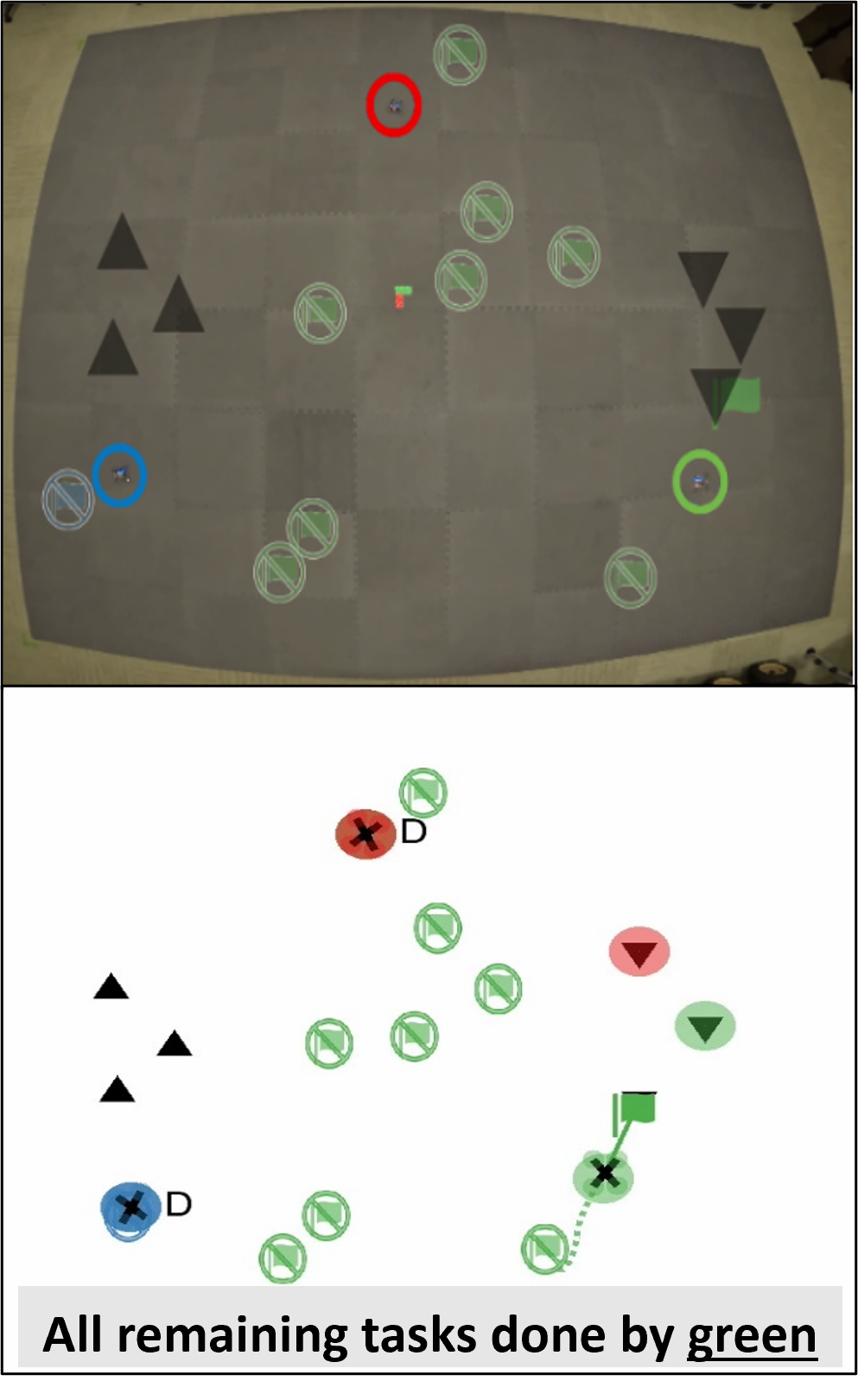}
    \label{fig:exf}
    }%
\vspace{-7pt}
\caption{Snapshots and results of an experimental case study.}
\label{fig:expFig}
\vspace{-17pt}
\end{figure*}

\section{Experiments}\label{sec:exps}
Our approach was validated through several laboratory experiments with a multi-robot team. The team consists of several Bitcraze Crazyflies that used a Vicon motion capture system for localization. 
Vehicles start within the communication range to complete all tasks in the environment. The experiments were carried out in a $4 \times 5.5$ [m] space with a sensing and communication range of $0.5$ [m] for each robot. The results of a sample experiment with ten tasks and three Crazyflies are shown in Fig.~\ref{fig:expFig}.

As shown in the figure, each robot is assigned a subset of tasks in Fig.~\ref{fig:exa}. After disconnection, the blue robot fails, and the green robot observes that the blue robot is not where expected in Fig.~\ref{fig:exb}; the green robot backtracks along the blue robot's path and finds the blue robot in Fig.~\ref{fig:exc}. In Fig.~\ref{fig:exd}, the green robot also observes that the red robot is not where expected. The green robot backtracks along the path of the red robot and finds the red robot in Fig.~\ref{fig:exe}. The green robot then replans all remaining tasks before ending at the depot in Fig.~\ref{fig:exf}.

\section{Conclusion}\label{sec:conclusion}
This paper presents a novel framework for multi-robot systems to use a modified centralized planning method to assign tasks, accounting for intermittent interactions. These interactions enable the system to use epistemic planning, which adapts to faults and disturbances by reassigning tasks based on a robot's reasoning about the system while disconnected. This method allows an MRS to disconnect and cooperatively plan based on a set of belief and empathy states if the system does not function as intended. The generalized task allocation algorithm uses these belief states to assign tasks while considering the potential need to communicate with disconnected robots, facilitating dynamic task allocation without constant communication. We show the improvement of our framework compared to a baseline heuristic over several scenarios and apply our framework to real-world experiments. 
Future research includes addressing the challenges of improving strategies for more complex environments. Additionally, we would like to reduce the computation time for task allocation and optimize the necessary belief propagation and interactions for a larger multi-robot system by dividing the team into sub-teams. Outdoor experiments are also on our agenda.

\section{Acknowledgements}
This work is based on research sponsored by: Northrop Grumman through the University Basic Research Program, the Swedish Research Council (VR) with the PSI project No. \#2020-05094, and the Knowledge Foundation (KKS) with the FIESTA project No. \#20190034.

\bibliographystyle{IEEEtran}
\bibliography{sample.bib}

\end{document}